\documentclass[journal]{IEEEtran}
\usepackage{booktabs}
\usepackage{ccsl}
\usepackage{times, mathptm}
\usepackage{multicol}
\usepackage{multirow}
\usepackage{setspace}
\usepackage{cite}

\usepackage{indentfirst}
\usepackage{array}
\usepackage[cmex10]{amsmath}
\usepackage{mdwmath}
\usepackage{mdwtab}
\usepackage[tight,footnotesize]{subfigure}
\usepackage{mdwlist}

\usepackage{subfigure}
\usepackage{caption}
\usepackage{yhmath}

\usepackage{moreverb}
\usepackage{graphicx}
\usepackage{mathrsfs}
\usepackage{bm}
\usepackage{amsmath}
\usepackage{amssymb}
\usepackage{supertabular}
\usepackage{threeparttable}
\usepackage{subfigure}
\usepackage{listings}

\usepackage{algorithm}
\usepackage{algorithmic}

\usepackage{yhmath}
\usepackage[amssymb]{SIunits}
\usepackage[colorlinks=false,
            linkcolor=red,
            anchorcolor=blue,
            citecolor=green,
            urlcolor=magenta
            ]{hyperref}

\DeclareMathSymbol{\mlq}{\mathord}{operators}{``}
\DeclareMathSymbol{\mrq}{\mathord}{operators}{`'}

\lstdefinelanguage{sketch} {
	morekeywords={
		harness, assert, void, minimize, checkCoincedence, checkPrecedence, IntersectionCnt, Intersection, UnionCnt,
        Union},
	sensitive=true,
	morecomment=[l]{//},
	morecomment=[s]{/*}{*/},
	morestring=[b]",
}

\def\figref#1{Figure~\ref{#1}}

\def\eqnref#1{Equation~(\ref{#1})}

\def\algoref#1{Algorithm~\ref{#1}}

\newcommand{\qod}{\nobreak \ifvmode \relax \else
      \ifdim\lastskip<1.5em \hskip-\lastskip
      \hskip1.5em plus0em minus0.5em \fi \nobreak
      \vrule height0.5em width0.5em depth0em\fi}

\begin{document}

\title{FDA$^3$: Federated Defense Against Adversarial Attacks for Cloud-Based IIoT Applications}

\author{Yunfei~Song, Tian~Liu, Tongquan~Wei, Xiangfeng~Wang, Zhe~Tao, and
 Mingsong Chen

\IEEEcompsocitemizethanks{
\IEEEcompsocthanksitem Yunfei Song, Tian Liu, Tongquan Wei, Xiangfeng Wang and Mingsong Chen
  are with the 
MoE Engineering Research Center of Software/Hardware Co-design Technology and Application, 
East China Normal University,
Shanghai, 200062, China (email: \{51174500122, 52184501017\}@stu.ecnu.edu.cn, \{xfwang, mschen\}@sei.ecnu.edu.cn, tqwei@cs.ecnu.edu.cn). 
Mingsong Chen is also
with the Shanghai Institute of Intelligent Science and Technology, Tongji
University.
Zhe Tao is with Godel Lab at Huawei, Shanghai, China (email:  taozhe1@huawei.com).
\IEEEcompsocthanksitem  
This article has been accepted for publication in a future issue of IEEE
Transactions on Industrial Informatics, but has not been fully edited. Content may change prior to final publication. Citation information: DOI 10.1109/TII.2020.3005969, IEEE
Transactions on Industrial Informatics.
\IEEEcompsocthanksitem  
Copyright $\copyright$ 2009 IEEE. Personal use of this material is permitted.
However, permission to use this material for any other purposes must be
obtained from the IEEE by sending a request to pubs-permissions@ieee.org.
See http://www.ieee.org/publications\_standards/publications/rights/index.html for more information.}}

\maketitle

\begin{abstract}
 Along with the proliferation of Artificial Intelligence (AI) and 
  Internet of Things (IoT) techniques, various kinds of adversarial attacks
are increasingly emerging to fool   Deep Neural 
  Networks (DNNs) used by Industrial IoT (IIoT) applications. 
  Due to biased training data or vulnerable underlying
 models, imperceptible  modifications
  on  inputs made by adversarial attacks may result in devastating consequences.
  Although existing  methods are promising in 
  defending such malicious attacks, 
  most of them can only deal with limited existing 
attack types, which
 makes  the deployment of 
  large-scale IIoT devices a great challenge.   
  To address this problem, we
  present an effective federated defense approach named $FDA^3$ that
  can aggregate
 defense   knowledge
 against adversarial examples from different sources.
 Inspired by federated learning, our proposed  cloud-based  architecture
  enables the sharing of 
  defense capabilities
 against different attacks among IIoT devices.
  Comprehensive experimental results show that the generated DNNs by our
 approach can not only 
  resist more malicious
 attacks  than existing attack-specific adversarial training
methods, but also can prevent IIoT applications from  new attacks.
\end{abstract}


\IEEEpeerreviewmaketitle

\section{Introduction}\label{sec:introduction}

Deep Learning (DL) techniques are increasingly deployed  in 
safety-critical   Cyber-Physical Systems (CPS) and Internet of Things 
(IoT) areas such as  autonomous driving,
commercial surveillance, and robotics,  where the prediction correctness 
of inputs is of crucial importance \cite{tii,tii1,dac1,deepxplore}.
However, along with the  prosperity of  Industrial
 IoT (IIoT) applications, 
they are inevitably becoming the main 
targets of malicious adversaries \cite{iotsurvey,tiiattack}.
No matter the adversarial attacks are intentional or unintentional, 
due to biased training data or  overfitting/underfitting models, 
the slightly modified inputs often make vulnerable 
IoT applications  demonstrate 
incorrect or unexpected behaviors,
which may cause disastrous consequences.



Most of existing adversarial attacks focus on generating  IIoT inputs with perturbations, which are named
``adversarial examples'' \cite{szegedy} to fool Deep Neural Networks (DNNs). Such adversarial examples can mislead the 
classifier models to predict
incorrect outputs, while they are not distinguishable by human eyes.     
To resist these attacks, various defense methods were proposed, e.g., ensemble diversity \cite{improvingadversarial},
PuVAE \cite{puvae}, and adversarial training. However, most of them are not suitable for 
IIoT applications. This is mainly because: i) most defense methods focus on defending one specific type of attacks; and 
ii) IIoT applications are usually  scattered  in different places
in face of  various adversarial
 attacks. 
In this situation, IIoT devices with the same type should be equipped
 with different DNNs to adapt  to different environments. 
Things become even worse when
various  new  adversarial attacks are
 emerging, since it is hard for IIoT designers
to quickly find a new solution to defend such attacks.


As a distributed machine learning approach, Federated Learning (FL)  \cite{communication} 
enables training of a high-quality centralized model over
a large quantity of decentralized data residing on IIoT devices. 
It has been widely studied
 to address the fundamental problems of privacy, ownership, and locality of data
for the cloud-based architecture, 
where the number of participating devices is huge but the  Internet connection is slow or 
unreliable. 
Based on the federated averaging technique
 \cite{towardsfederated}, FL
 allows
 the training of a DNN without revealing the data stored on IIoT devices. The weights of 
new DNNs are synthesized using FL in the cloud, constructing a global model which is then pushed back 
to different IIoT devices for inference.  However, so far none of  existing FL-based approaches 
investigated  the defense of adversarial attacks for
 IIoT applications.


Since cloud-based architectures can extend  processing
capabilities of IoT devices by offloading  their 
partial computation tasks to remote cloud
servers,   the combination of cloud computing and IoT  
is becoming a popular paradigm 
that enables  
large-scale  intelligent IIoT applicationsm, where
IIoT devices  are connected with the cloud in a CPS
context  \cite{edge1}. 
However, no matter whether  the role of  cloud servers is 
for training or inference, IIoT devices are required to 
send original data to cloud servers, where the network latency  and  data 
privacy issues  cannot be neglected. 
Moreover,  if  devices of an IIoT application adopt  DNNs with the same type, the 
robustness of the application can be easily violated due to the varieties of adversarial 
attacks. Therefore, how to generate
 a robust DNN for a large quantity of 
IIoT devices with the same type while the privacy of these devices can be protected is becoming a 
challenge. Inspired by the concept
 of federated learning, 
this paper presents an effective federated defense framework named $FDA^3$ 
for large-scale IIoT applications. 
It makes   following {\bf three major contributions}:
\begin{enumerate}
\item We propose a new loss function for the  adversarial training on IIoT devices, which 
fully takes  the diversities of adversarial attacks into account. 
\item We present an 
efficient federated adversarial learning scheme that  can 
derive robust
 DNNs to resist
 a wide spectrum of adversarial attacks.  
\item  Based on the cloud-based 
architecture, we introduce a novel federated defense framework  for 
large-scale IIoT applications. 
\end{enumerate}
Experimental results on two well-known benchmark
 datasets  show that the DNNs generated by our proposed approach on IIoT devices
can resist more adversarial attacks than state-of-the-art methods. Moreover, 
the robustness of the generated DNNs  
 becomes better when the size of 
investigated IIoT applications grows larger.

The rest of this paper is organized as follows.
Section~\ref{sec:related} introduces the related work on 
defense mechanisms against adversarial attacks for DNN-based IoT designs.
Section~\ref{sec:framework} 
introduces
our federated defense framework  $FDA^3$ in detail.
 Section~\ref{sec:experi} presents experimental results, showing the effectiveness and 
 scalability of our approach. Finally, 
Section~\ref{sec:conclusions} concludes the paper.

\section{Related Work}\label{sec:related}

When more and more safety-critical IoT applications adopt DNNs, 
the robustness of DNNs is becoming a major concern 
in IoT design 
 \cite{deeplearningiot,deepxplore,suting1}.
The vulnerability of DNNs has been widely investigated by various malicious
 adversaries, 
who can generate physical adversarial examples to fool DNNs \cite{adversarial,jiliang}. 
Typically, existing attack methods can be classified into two categories, i.e., 
white-box attacks which assume that  DNN
structures are available, and black-box attacks without the knowledge of DNN architectures.
For example, as a kind of well-known white-box attack, 
Fast Gradient Sign Method (FGSM) \cite{expaliningandharnessing}
tries to add adversarial 
 perturbations in the  direction of the loss gradients.
In \cite{adversarial},  Kurakin et al. introduced the 
Basic Iterative Method (BIM) by applying FGSM  multiple times with small step size. 
In \cite{thelimitations}, the Jacobian-based Saliency Map 
Attack (JSMA) is introduced to identify features of the
input that most significantly impact output classification. JSMA 
crafts adversarial examples based
on computing forward derivatives.
To minimize the disturbance while achieving  better attack effects, 
Carlini and Wagner \cite{towardsevaluation} designed an optimization function-based attack, named CW2.
In \cite{deepfool}, DeepFool is proposed that uses  iterative linearization and geometric formulas to generate 
 adversarial examples. Unlike above attacks, SIMBA \cite{simpleblack} is a kind of
 simple but effective black-box attack.
Instead of investigating gradient directions as 
used in FGSM, SIMBA picks random directions to perturb images.

To address adversarial
 attacks, various defense mechanisms have been investigated. 
Typically, they can be classified into three categories \cite{puvae}. The first type (e.g., ensemble diversity 
\cite{improvingadversarial}, Jacobian Regularization \cite{improvingdnn}) optimizes
the  gradient calculation of  target classifiers. However, when processing nature images,
the performance of these defense approaches may degrade.     
The second type (e.g., PuVAE~\cite{puvae}, feature squeezing~\cite{featuresqueezing})
tries to purify the inputs of target classifiers via extra auto-encoders or filters. 
Nonetheless, the extra facilities inevitably increase the workload of host IIoT devices. 
By modifying training data, the third type can be used to regularize target classifiers.
As an outstanding example of the third type, adversarial training
 \cite{expaliningandharnessing,ensembleadversarial}
targets to achieve robust classifiers based on the training on both adversarial and
 nature examples. 
However, existing 
adversarial training methods for  specific  
IIoT devices only focus on a limited number of
 adversarial attack types. Therefore,
 the trained models using above methods usually 
cannot be directly used by devices deployed in new environment.

Aiming at improving the performance of training,
federated learning \cite{konevcny2016federated,communication}
is becoming widely used in generating
 centralized models while ensuring high data privacy for 
large-scale distributed applications \cite{towardsfederated}.
Although it has been studied in IIoT design, it focuses
 on 
the data privacy issues \cite{iotfla} rather than adversarial attacks. 
To the best of our knowledge, our work is the first attempt that adopts 
the concept of federated learning to construct a defense framework against 
various adversarial attacks for large-scale IIoT applications.

\section{Our Federated Defense Approach}\label{sec:framework}

Due to  privacy information leakage,
 adversaries can obtain 
 DNN model information from IIoT devices
 for attack purposes. In our approach, under the help of privacy 
protection mechanisms provided by IIoT devices,
 we assume that  the model cracking time is 
longer than the model update period. 
In this case, adversaries
 always cannot obtain the latest version of models used by IIoT devices.
However, adversaries
 can use  transfer attacks  to fool DNN models 
based on the privacy information they obtained. 
This paper focuses on how to 
 retrain the threaten
 model, so that it can resist various types of adversarial examples generated by adversaries. 
The following subsections
will introduce 
 our cloud-based federated defense  architecture, loss function  for device-level federated adversarial training, and  model update and synchronization
 processes in detail.

\subsection{The Architecture of $FDA^3$}

\begin{figure*}[htbp]
	\centering
	\includegraphics[width=6.5in]{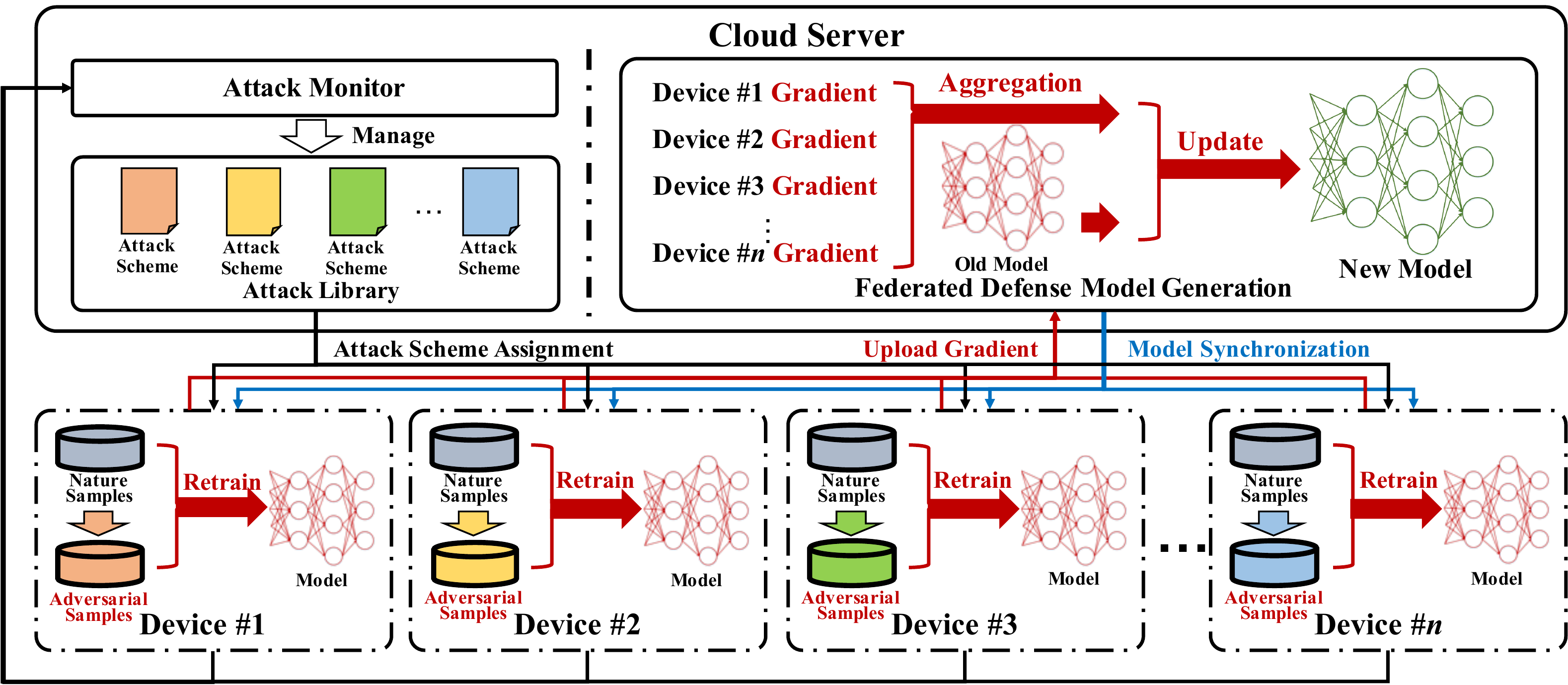}
	\caption{The framework and workflow of $FDA^3$}\label{fig:framework}
\end{figure*}

Figure~\ref{fig:framework} details the framework of $FDA^3$ together with its workflow,
 which  is inspired by the adversarial training  and federated learning methods.
The architecture of our approach consists of two parts, i.e.,   IIoT devices and their 
cloud server.  Besides the function of inference, the DNNs resided in 
IIoT devices  are  responsible for the DNN evolution for resisting  adversarial examples. 
Initially, all the devices
 share the same DNN. Since they are deployed in different environments, 
they may encounter  different input examples  and   different types of attacks. 
Such imbalances make the federated learning a best solution to aggregate different defense capabilities.

The cloud server consists
 of two modules, i.e., attack monitor module
 and federated defense model 
generation module. The attack monitor module is used to record the latest attack information 
for IIoT devices according to their locations or types. 
The module manages a  library consisting of all the reported attack schemes (i.e., source code or 
executable programs).  Such 
information can be collected by  IIoT device producers or from third-party  institutions. When 
the attack monitor module  detects some new attack for a device,  
 it will 
require
 the  device to download the corresponding
 attack scheme  for the purpose of adversarial
 retraining. Similar to 
 federated learning, the federated defense model generation module periodically collects  
device gradient information  and aggregates them to achieve an updated model with better 
robustness. Then the module will dispatch the newly formed model to all the  connected IIoT 
devices for the purpose of
 model synchronization.

During the execution of an IIoT device, the device
 keeps a buffer to hold a set of 
nature examples that are collected randomly
when their prediction confidence is high.  
For a specific period, all the IIoT devices need to be retrained and synchronized 
in a federated learning way. This process consists of three steps. First, based on the 
assigned attack schemes  by the cloud server, 
each device generates  corresponding adversarial examples locally to form
 a retraining set, whose elements are pairs of nature examples and 
corresponding adversarial examples. In the second step, the local adversarial
 training 
process 
 periodically
uploads the newly achieved  gradient information from
  IIoT devices
 to the cloud server for the model  
update and  synchronization. Finally, similar to federated learning,
the model generated  by our federated defense approach will be deployed on each
connected IIoT devices. Note that when a new IoT devices joins the IIoT application, it  needs to download the new model from the server. 
Due to the diversities of different devices, the 
new model is more robust, which  can resist
more
 attacks of different types. Since the interactions between 
the cloud server and IIoT devices only involves gradient information, the 
data privacy 
of IIoT devices can be guaranteed.


\subsection{Loss Function Modeling}\label{sec:gradient-computing}

In machine learning, loss functions are 
used for classification representing the 
cost paid for inaccurate predictions. 
In the context of deep learning, the loss functions can be used in the training phase
 to optimize the 
model prediction accuracy. 
When taking adversarial attacks into account, the definitions of loss functions for 
adversarial training are different from traditional ones.
Most existing loss functions for adversarial training 
consist of two components, i.e., the normal loss function and the adversarial loss function 
\cite{expaliningandharnessing}.   They can be formulated as:
\begin{equation}
{\mathcal{L}} \left(x,\hat{x},y|\theta \right)= \left(1-\alpha \right) {\mathcal{L}}_{normal} \left(x,y \mid \theta \right) + \alpha {\mathcal{L}}_{adv} \left(\hat{x},y \mid \theta \right) 
\label{equ:1}.
\end{equation}
The notation ${\mathcal{L}}_{normal} \left(x,y \mid \theta \right)$ 
denotes the normal loss function, where
 $x$ and $y$ represent the nature (normal) example and the classification label when the model's parameter is $\theta$. 
Similar to the definition of ${\mathcal{L}}_{normal} \left(x,y \mid \theta \right)$, 
the notation 
${\mathcal{L}_{adv}} \left( \hat{x} , y \mid \theta \right)$ is used to 
denote the adversarial loss function, where $\hat{x}$ denote the adversarial example generated from 
$x$. 
The notation ${\mathcal{L}} \left(x, \hat{x} , y \mid \theta \right)$ represents the 
overall loss function of  adversarial training.
The hyperparameter $\alpha$ ($\alpha \in \left[ 0,1\right]$)
in Equation~(\ref{equ:1}) is used to set
 the proportion of the adversarial loss function in the overall loss function. The higher 
 value of $\alpha$ indicates the higher weight of the  
adversarial loss function  contributing
 to the overall loss function.

For different adversarial attacks, there exist different ways to achieve an optimal 
$\theta$  that can   minimize 
${\mathcal{L}} \left(x, \hat{x}, y \mid \theta \right)$. As an example
 of FGSM attacks which are of $L_{\infty}$ norm,  the optimal
 $\theta$ can be calculated using: 
\begin{equation}
\theta^* =  \mathop{\arg\min}_{\theta}  \mathop{\mathbb{E}}_{x\subseteq D} \left[ \mathop{max}_{ \|\hat{x} - x \|_{\infty} \leq \epsilon}  {\mathcal{L}} \left(x,\hat{x},y\mid\theta \right) \right]
\label{equ:2},
\end{equation}
 where $D$ denotes the training set and the  $L_{\infty}$ norm  $\|\hat{x} - x \|_{\infty} \leq \epsilon$ specifies the 
allowable difference  between nature and adversarial examples. 
Note that \eqnref{equ:2} focuses on adversarial attacks of $L_{\infty}$ norm (e.g., FGSM, BIM).
It cannot  deal with adversarial attacks of $L_{0}$ norm (e.g., JSMA) and $L_{2}$ norm (e.g., CW2, DeepFool, SIMBA).
To cover  all  possible attacks for an IIoT application with
 $N$ devices,  we extend ${\mathcal{L}} \left(x,\hat{x},y|\theta \right)$ defined in 
\eqnref{equ:1} as follows: 
\begin{equation}
\begin{split}
{\mathcal{L}}_{fed} & \left(x,\hat{x},y \mid \theta \right)=  \frac{1}{N} \Sigma_{k=1}^N {\mathcal{L}} \left(x^k,\hat{x}^k,y^k \mid \theta \right).
\end{split}
\label{equ:3}
\end{equation}
The notation ${\mathcal{L}}_{fed} \left(x,\hat{x},y \mid \theta \right)$  denotes 
the loss function of our federated defense approach, which equals to the 
arithmetic average of all the loss functions of $N$ devices participating in  adversarial training. Here,  $\hat{x}$ denotes the set of all the adversarial examples generated
 by the $N$ devices, and $x$ and $y$ denote the nature examples and 
their corresponding labels, respectively.
The symbols $x^k$, $\hat{x}^k$, $y^k$ denote the nature examples, adversarial examples and 
corresponding labels of the $k^{th}$ device. 
The optimization target of our federated adversarial training 
is to figure out   an optimal
 $\theta$ which can be formulated as:
\begin{equation}
\begin{split}
\theta^* = \mathop{\arg\min}_{\theta} \Bigg\{&  \mathop{\mathbb{E}}_{x\subseteq D,\hat{x} \subseteq D_{L_{\infty}}} \big[ \mathop{max}_{||\hat{x} -x ||_{\infty} \leq \epsilon}  {\mathcal{L}}_{fed}\left(x,\hat{x},y|\theta\right)\big] +\\
&\mathop{\mathbb{E}}_{x\subseteq D,\hat{x} \subseteq D_{L_{0}}}  \big[ \mathop{max}_{||\hat{x} -x ||_0 \leq \sigma}  {\mathcal{L}}_{fed}\left(x,\hat{x},y|\theta\right) \big]  +\\
& \mathop{\mathbb{E}}_{x\subseteq D,  \hat{x} \subseteq D_{L_{2}}} \big[ \mathop{max}_{||\hat{x} -x ||_2 \leq \delta}  {\mathcal{L}}_{fed}\left(x,\hat{x},y|\theta\right) \big] \Bigg\}
\end{split}
\label{equ:5}
\end{equation}

In \eqnref{equ:5} we use $D_{L_{0}}$, $D_{L_{2}}$
and $D_{L_{\infty}}$ to indicate generated adversarial examples of norm $L_{0}$, norm $L_{2}$ and 
$L_{\infty}$ from $D$, respectively.
We can find that \eqnref{equ:5} tries to figure out one comprehensive 
 defense model that can resist 
a wide range of  known attacks  with higher accuracy than locally retrained models.


\subsection{Federated Defense Model Generation}\label{sec:parameter-updates}

Our federated defense approach consists of two parts, i.e.,  IIoT devices 
and corresponding
 cloud server. During the  execution, IIoT devices randomly
 collect a set of  nature examples 
 with high confidence on-the-fly
 and save them in their local memory. Based on   attack schemes assigned by the cloud 
sever, IIoT devices  generate adversarial examples for their model
 re-training.  
Note that due to the limited resources (e.g., memory size, computing power)
 of IIoT devices, usually cloud servers only assign  limited number of attack schemes
to IIoT devices. 
Similar to federated learning, the adversarial training process of our federated defense method
involves multiple epochs, where 
an epoch may involve multiple iterations based on the user specified batch size.
In our approach, we consider each iteration as a round. Within a round,
 all the IIoT devices  send  gradient information obtained from their 
locally retrained models to the cloud server, and then the cloud server aggregates the 
gradients and synchronizes the updated model with all the IIoT devices.

{
\begin{algorithm}
	\renewcommand{\algorithmicrequire}{\textbf{Input:}}
	\renewcommand{\algorithmicensure}{\textbf{Output:}}
	\caption{Adversarial Training Procedure for IIoT Devices}
	\label{alg:2}
	\begin{algorithmic}[1]
		\REQUIRE i) $S$, cloud server; \qquad   ii) $att$, adversarial attack types; \\
		\quad iii) $b$, batch size;  \qquad \ iv) $\alpha$,  hyperparameter; \\
		\quad \ v) $E$, \# of epochs;  \ \ \ \ vi) $K$, device index; \\
	\ \  vii) $\mathbb{F}$, device model;
     
		\ENSURE 
		

\WHILE{$true$}
		\STATE $x_{nat} = Collect\left(\right) $
		\STATE $x_{adv} = AdvGen \left( x_{nat}, att, \mathbb{F}, \theta^K\right) $ \quad $\triangleright$  $\theta^K$, device model weight 

		\STATE $y_{nat} = \mathbb{F} \left( x_{nat},\theta^K \right) $ 

		\STATE $y_{adv} = \mathbb{F} \left( x_{adv},\theta^K \right) $
		\STATE $x_{nat}^A = Augumentation\left(x_{nat}, |att|\right)$
		\STATE $y_{nat}^A = Augumentation\left(y_{nat}, |att|\right)$

\FOR {$e \in \left\{1, 2, \cdots, E\right\}$} 
		\FOR {$i \in \{1,2,\cdots,\lceil\frac{|x_{nat}|}{b}\rceil\}$}

		\STATE $\left(  x_{nat}^{A,i},y_{nat}^{A,i},y_{adv}^{i} \right)= partition\left(x_{nat}^A,y_{nat}^A,y_{adv},b,i\right)$
                  
                \STATE $y_{nat}^{A,i*}=\mathbb{F} \left( x_{nat}^{A,i},\theta^K \right)$
 		\STATE $y_{adv}^{i*}=\mathbb{F} \left( x_{adv}^{i},\theta^K \right)$
                 
		\STATE $normal\_loss = CrossEntropy \left( y_{nat}^{A,i},y_{nat}^{A,i*} \right) $
		\STATE $adv\_loss = CrossEntropy \left( y_{nat}^{A,i},y_{adv}^{i*} \right) $
		\STATE $Loss = \alpha * normal\_loss + \left( 1-\alpha \right) * adv\_loss$
		\STATE $\nabla \theta^{e, i} = AdamOptimizer.compute\_gradients \left( Loss,\theta^K \right)$

		\STATE $send \left( \nabla \theta^{e, i},K, S \right) $  \ $\triangleright$  Send gradients  to cloud server

		\STATE $\theta^K=sync\_recv\left(K,S\right)$ 
		
\ENDFOR
\ENDFOR
		
\ENDWHILE

	\end{algorithmic}  
\end{algorithm}}

\algoref{alg:2} details the local adversarial training procedure for IIoT devices. 
Note that we assume that the IIoT device with index $K$  has been connected to the server and its model is the same 
as the one of the server initially. 
In step 2, the device 
tries to collect nature examples with high prediction confidence randomly. 
Based on the old model (i.e., $\mathbb{F}$ and $ \theta^K$) and
 the assigned attack schemes $att$ by the cloud server, 
step 3 generates the adversarial examples for $X_{nat}$ using  transfer attacks 
attacks. Note that
 if the cardinality
of    $att$ (i.e.,  $|att|$) is larger than one, step 3 will generate  $|att|$
adversarial attacks of
 different types for each example
 in $X_{nat}$. 
Since all the examples in $X_{nat}$ are collected with high confidence, 
step 4 tries to figure out
 the  labels for them. 
Similar to step 4, step 5 obtains
 the prediction results for 
all the generated
 adversarial examples. Step 6 enlarges  both $X_{nat}$ and $Y_{nat}$
by duplicating them $|att|$ times for the following loss function 
calculation in steps 13-14. Steps 8-20 iteratively
interact with the cloud server, where 
steps 10-18 form one round for the gradient aggregation and model update.
Step 10 divides the nature and adversarial examples batch by batch. 
Steps 11-12 figure out the prediction labels for nature and adversarial 
examples in the same batch, respectively. Steps 13-15 calculate the  overall 
loss function based on
  the nature and adversarial examples in the batch 
using the equation defined in \eqnref{equ:1}.  
Step 16 
computes the gradient information and step 17 sends it to the cloud server. 
Step 18 updates the local model  using the aggregated
 gradient information sent by 
the cloud server. Note that the  $sync\_recv$ is a blocking function 
waiting for the reply from the cloud server.

{
\begin{algorithm}
	\renewcommand{\algorithmicrequire}{\textbf{Input:}}
	\renewcommand{\algorithmicensure}{\textbf{Output:}}
	\caption{Model Generation Procedure for Cloud Server}
	\label{alg:3}
	\begin{algorithmic}[1]
		\REQUIRE 
		  i) $\theta_S$,  weight of the server model; \\
                 \quad \  ii) $b$, batch size;  \\
		\ \  \  iii) $E$, \# of epochs; \\
                 \  \  \    iv) $N$, \# of devices;\\
                \ \ \ \ v) $nat$, \# of nature examples on one device;
		\ENSURE 
\WHILE{$true$}
		\FOR {$ e \in \left\{1, 2, \cdots, E\right\}$}
		\FOR {$i \in \{1,2,\cdots,\lceil\frac{|nat|}{b}\rceil\}\}$} 
		\FOR {$ K \in \{1,2,\cdots,N\}$}
		 \STATE $\nabla \theta_K^{e, i} =receive\left(K\right)$   \quad $\triangleright$ Receive $K^{th}$ device's gradients
		\ENDFOR
		\STATE $\nabla \theta_S^{e, i} = \frac{1}{N} \Sigma_{K=1}^N \nabla \theta_K^{e, i} $  \quad $\triangleright$  Gradient aggregation

		\STATE $\theta_S = AdamOptimizer.apply\_gradients \left( \theta_S,\nabla \theta_S^{e, i} \right) $ 
		\FOR {$ K \in \{1,2,\cdots,N\}$}
		\STATE $sync\_send \left(K, \theta_S \right) $ 
		\ENDFOR
		\ENDFOR
		\ENDFOR

\ENDWHILE

	\end{algorithmic}  

\end{algorithm}
}

\algoref{alg:3} presents the model generation procedure conducted by the cloud server involving 
both model aggregation and model update operations. As shown in steps 2-13, the sever needs to 
perform $E\times \lceil\frac{|nat|}{b}\rceil$ rounds of  interactions with all the devices to form a 
robust model for federated defense. After receiving  gradients from 
all the devices in step 5, step 7 aggregates the gradients according to \eqnref{equ:3}, and step 8
applies
 the aggregation result to the current model, i.e., $\theta_S$. Note that the 
$receive$ function in step 5 is a blocking function. When one round is finished, steps 9-11
send the newly updated model weight information to each connected IIoT devices for the 
model
synchronization.

\section{Experiments}\label{sec:experi}

To evaluate the effectiveness
 of our approach, we implemented our $FDA^3$ approach
on top of a  cloud-based  architecture, which consists of a cloud server and 
a set of connected IIoT devices. Since we focus on  classification accuracy, the behavior of cloud servers and IIoT devices
are all simulated on a workstation
 with Intel  i7-9700k CPU,  16GB memory,  NVIDA GeForce GTX1080Ti GPU, and 
Ubuntu operating system.  
We adopted Tensorflow (version 1.12.0) and  Keras (version 2.2.4) to construct all DNN models in our framework. 
To validate our approach, we conducted two case studies
 using LeNet \cite{lenet} for dataset 
MNIST \cite{mnist}, and ResNet \cite{resnet} for CIFAR10 \cite{cifar10}, respectively.
The initial  LeNet model is 
 trained using 60000 training
examples from the MNIST dataset, and the initial 
ResNet model is trained using 50000 training examples from the CIFAR10 dataset.
Note that 
both MNIST and CIFAR10 datasets have a test set of 10000 examples, individually. We 
divided them  into two halves, where 5000 examples are used for re-training and the remaining 5000
examples are used for testing.



\subsection{Performance Comparison}

In the first experiment, we considered
 an  IIoT application that has 10 devices  connected to 
the same cloud server for federated defense. Since most IIoT devices are 
memory constrained, we assumed that the DNN model on an IIoT device can
 only keep 100 nature examples for adversarial training.
We considered five well-known types of  attacks in the 
experiment, i.e., FGSM \cite{expaliningandharnessing},
BIM \cite{adversarial}, JSMA \cite{thelimitations}, CW2 \cite{towardsevaluation}, 
and DeepFool \cite{deepfool}, where each type was
 used to attack two out of the ten
 IIoT devices. 
To enable adversarial training,  for each device we 
generated 100 adversarial examples for the 100 nature
 examples using the  assigned attack scheme,
 respectively.
Note that all the adversarial examples here were generated by transfer attacks, assuming that 
the initial model can be obtained while the intermediate
retrained models cannot be accessed by 
malicious adversaries. 
Similar to   \cite{expaliningandharnessing}, we set the 
 hyperparameter $\alpha$ to 0.5, which indicates that both normal and adversarial losses  contribute to  the total loss equally. 

For the federated training of adversarial attacks, we set the batch size to 
100 pairs of nature and adversarial examples. In this case, an epoch consists of only 
one iteration, which can be considered as a round
for the retraining of  all the collected  example pairs on a device. We set the 
epoch number to 50.
Once one epoch  is finished, the IIoT devices need to send the updated 
gradient information to their corresponding
 cloud server to perform  aggregation.

\begin{figure}[htbp]
	\centering
	\includegraphics[width=1\linewidth]{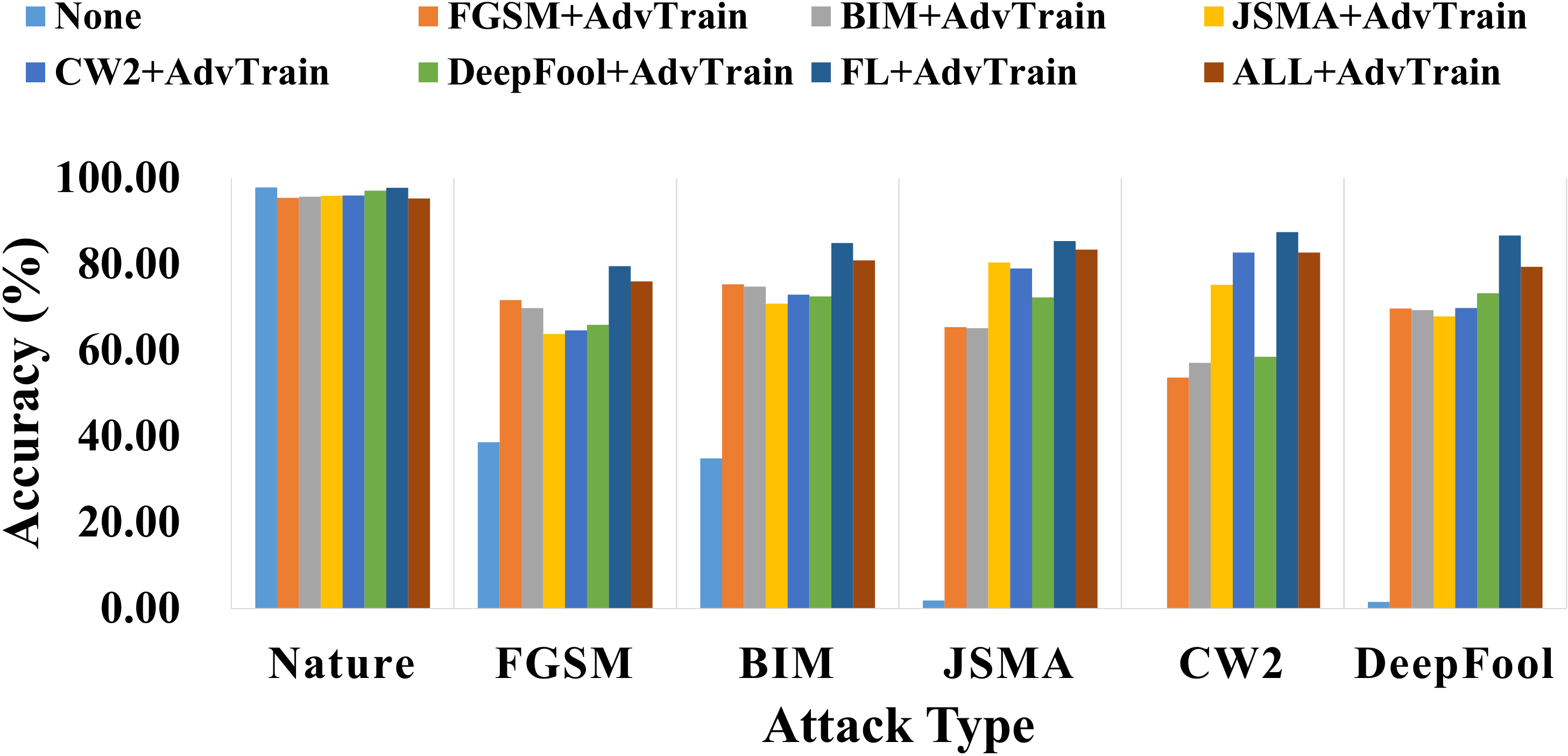}
	\caption{Performance comparison between different defense methods for MNIST dataset}
	\label{fig:compare-mnist}
\end{figure}

To enable the performance comparison with the model derived by our federated defense approach,
we also generated the models by training  the 100 example pairs locally using different 
types of attacks individually.  
We use the notation {\it None} to denote the initial model without any retraining. 
 We use the notation {\it X+AdvTrain} ({\it X}$\in \{FGSM,BIM,JSMA,CW2, DeepFool\}$) to indicate the model of an IIoT device that
is retrained locally based on the 100 adversarial examples made by  attacks of type {\it X}. 
The notation {\it FL+AdvTrain} denotes the model generated by our federated 
defense approach. For fair comparison, we also applied all the five attack types on a randomly 
selected IIoT 
device among the 10 devices.
 In this case, we generated 500 adversarial examples to retrain the model  
locally (with a batch of 100 example pairs) and got
 the new model {\it ALL+AdvTrain}. 
Figure~\ref{fig:compare-mnist} shows the inference accuracy results for 
the MNIST dataset. As shown on the X-axis, we used the notation 
{\it Nature} to denote the 5000 test examples without any attack from the MNIST dataset.
The other notations on X-axis indicate
 the adversarial test   sets generated  by a specific attack type based on {\it Nature}.

From Figure~\ref{fig:compare-mnist}, we can find that  {\it FL+AdvTrain}
achieves the best prediction accuracy among seven out of eight
  models except  {\it Nature}.
For the  {\it Nature} test set, the {\it None} method slightly outperforms 
our approach by 0.08\%.
This is because  the {\it FL+AdvTrain} model
includes new  adversarial examples in the retraining.
For the remaining 
five test sets, we can find that  {\it FL+AdvTrain}  outperforms
the other seven models significantly. 
As an example of 
DeepFool test
 set, the {\it FL+AdvTrain} model can achieve better 
accuracy than the {\it DeepFool+AdvTrain} model by 13.5\%, though {\it DeepFool+AdvTrain} is retrained 
specific for  DeepFool attacks.  Note that comparing with the  {\it ALL+AdvTrain}
model,  our {\it FL+AdvTrain} shows much better accuracy for all the six attack types. 
In other words, our federated learning is a better choice for IIoT deployment than all the 
locally retrained models for specific devices.

\begin{figure}[htbp]
	\centering
	\includegraphics[width=1\linewidth]{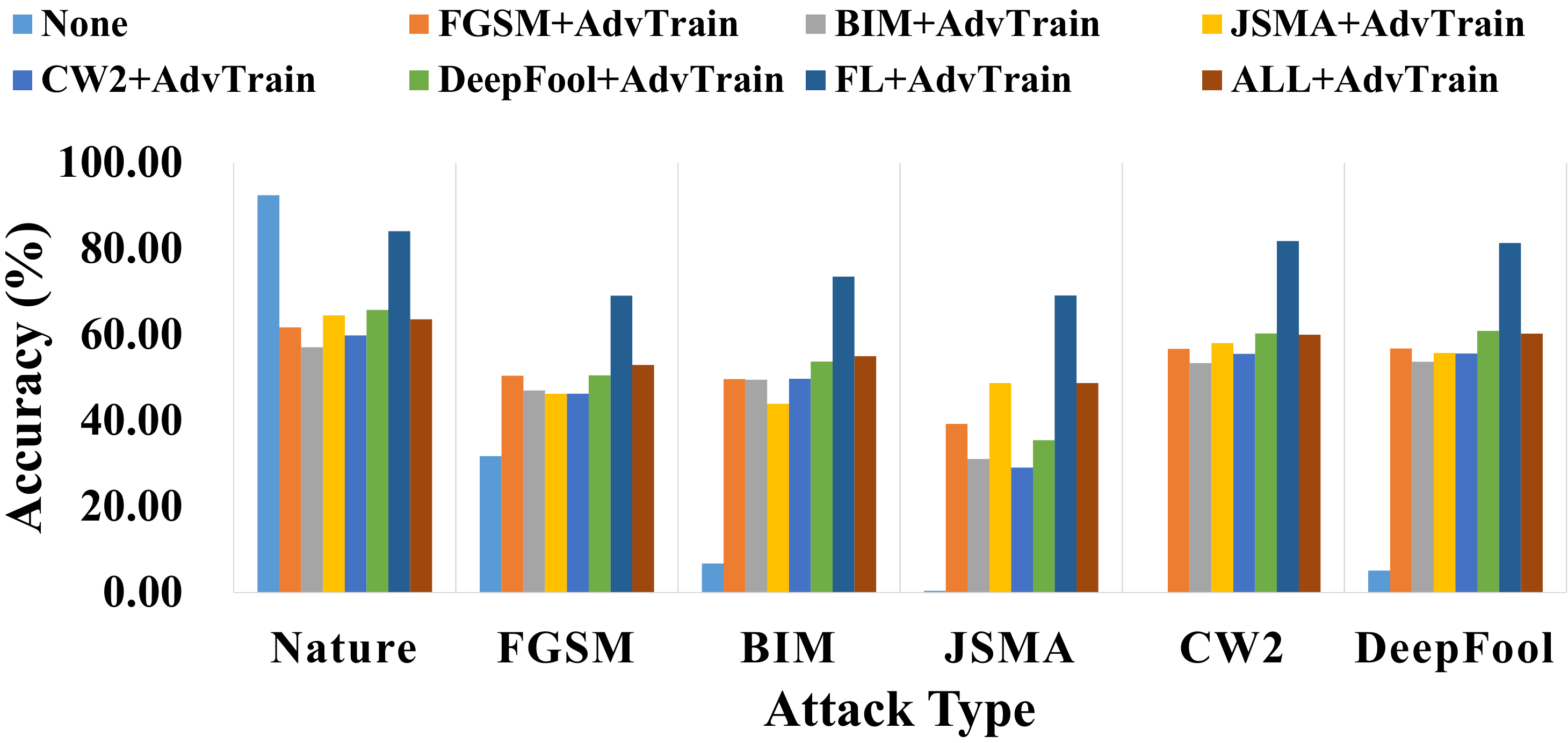}
\caption{Performance comparison between different defense methods for CIFAR10 dataset}
	\label{fig:compare-cifar10}
\end{figure}

We also checked the performance of federated defense method on the CIFAR10 dataset. 
Figure~\ref{fig:compare-cifar10} shows the comparison results. 
Similar to the observations from Figure~\ref{fig:compare-mnist}, we can find that 
our approach outperforms the other seven methods. In this experiment, 
we can observe that the {\it ALL+AdvTrain}
model  shows better accuracy comparing with other models with specific attacks. 
 	
\begin{figure}[htbp]
	\centering
	\includegraphics[width=1\linewidth]{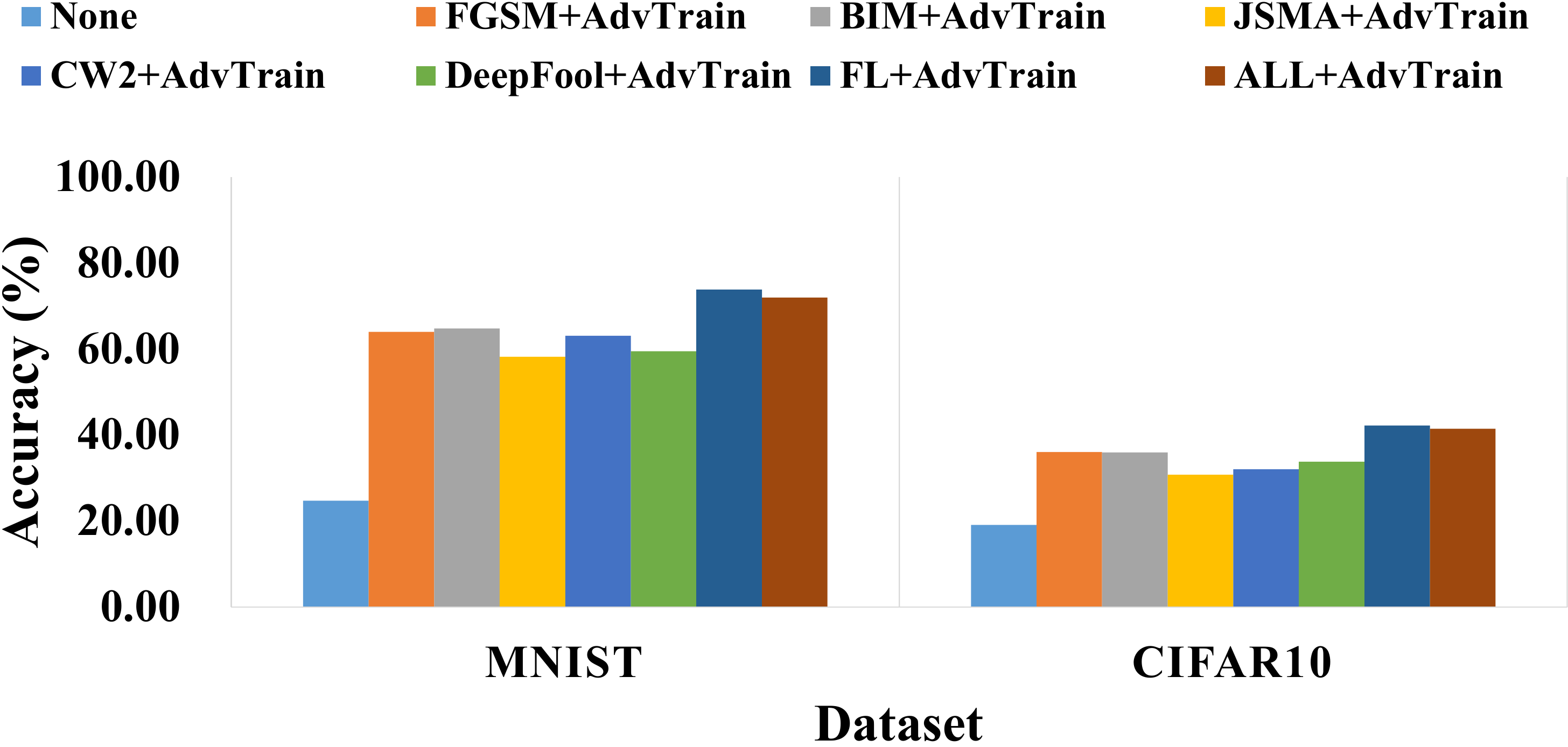}
	\caption{Performance comparison between different defense methods for SIMBA attack}
	\label{fig:simba-vs}
\end{figure}

Since IIoT devices are deployed in an uncertain environment with more and more emerging 
new attacks, we also checked the robustness
 of models generated by our approach using new attacks. \figref{fig:simba-vs} presents
 the
robustness comparison results
 between different defense methods against 
a new type of attacks, i.e., SIMBA
 \cite{simpleblack}.  We  used the eight
 models generated in  
 \figref{fig:compare-mnist} and \figref{fig:compare-cifar10} for this new attack. 
We generated 5000 test examples using SIMBA and applied each model on this test set
individually. We can find that for both MNIST and CIFAR10 datasets our {\it FL+AdvTrain}
model can resist more SIMBA attacks than the 
other models. As an example for MNIST test set, 
our method has an accuracy of 73.8\%, while the  {\it ALL+AdvTrain}
model only has an  accuracy of 71.9\%.
 In other words, the robustness of our 
federated defense method is  better than other local adversarial training methods to defend
new attacks.

\subsection{Scalability Analysis} 

\begin{figure}[htbp]
	\centering
	\includegraphics[width=1\linewidth]{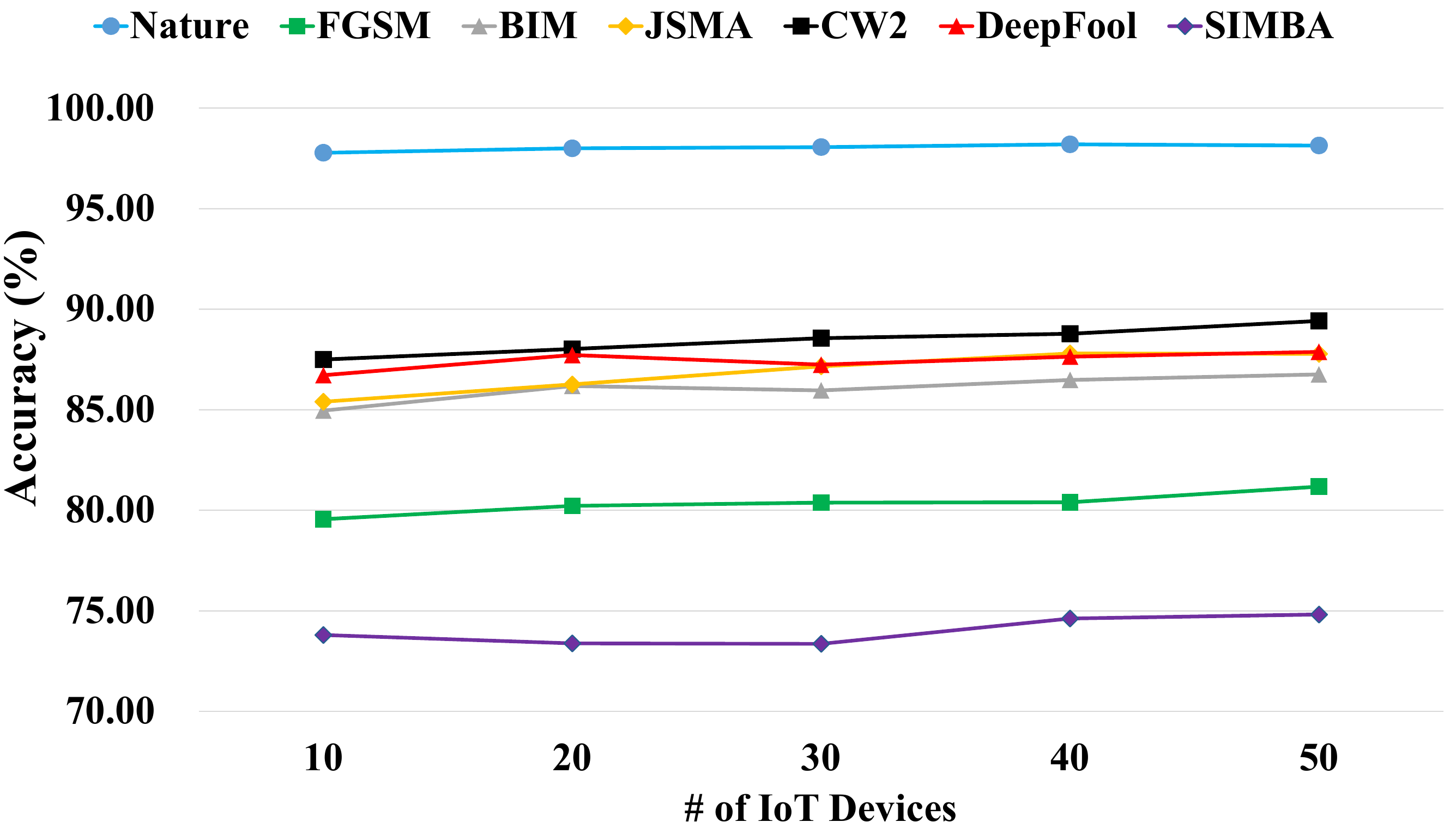}
	\caption{The impact of the number of IIoT devices for our federated defense methods on  
MNIST dataset}
	\label{fig:number-node-mnist}
\end{figure}

The first experiment only investigates an IIoT application with only 10 devices.
However,  a typical IIoT application may involve dozes of or hundreds of 
devices. Therefore, to verify whether our approach can be applied to
 large-scale
IIoT applications, we conducted the second experiment to check the 
 scalability of our approach. 
\figref{fig:number-node-mnist} shows the trend of prediction accuracy along with the 
increment of the number of IIoT devices over the MNIST dataset.
In this experiment, we used our $FDA^3$ to generate the model for IIoT devices. 
Similar to the scheme used in the first experiment, we considered
 five attack types
(i.e., FGSM, BIM, JSMA, CW2, DeepFool) for the adversarial example generation. We assumed that 
there were one fifth of devices  attacked   by a specific type of attacks. 
For example, if there are 10 devices involved in an IIoT application, there will be 2 devices for 
each of the five investigated attack types. 
 We investigated 7 types of test examples,
where {\it Nature} denotes a test set of 5000 nature examples, and the other six 
test sets of 5000 examples each are labeled using the attack type.

From \figref{fig:number-node-mnist} we can find that the prediction accuracy for {\it Nature}
is the highest. 
Moreover, we can find that when more devices are engaged in federated defense, the 
accuracy for {\it Nature} test set can still be slightly
 improved.  
The same trend can be  observed from
 the other six adversarial test sets.
As an example for the JSMA test set, when the number of IIoT devices increases from 10 to 50, 
the accuracy can be improved from 85.4\% to 87.8\%.
Note that in the federated defense, the attack type SIMBA  is not considered. Therefore, 
we can observe that  the prediction accuracy for  SIMBA adversarial test 
set is the lowest. However, we can find an accuracy improvement along the increment of 
the number of devices, i.e., from 73.8\% to 74.8\%.

\begin{figure}[htbp]
	\centering
	\includegraphics[width=1\linewidth]{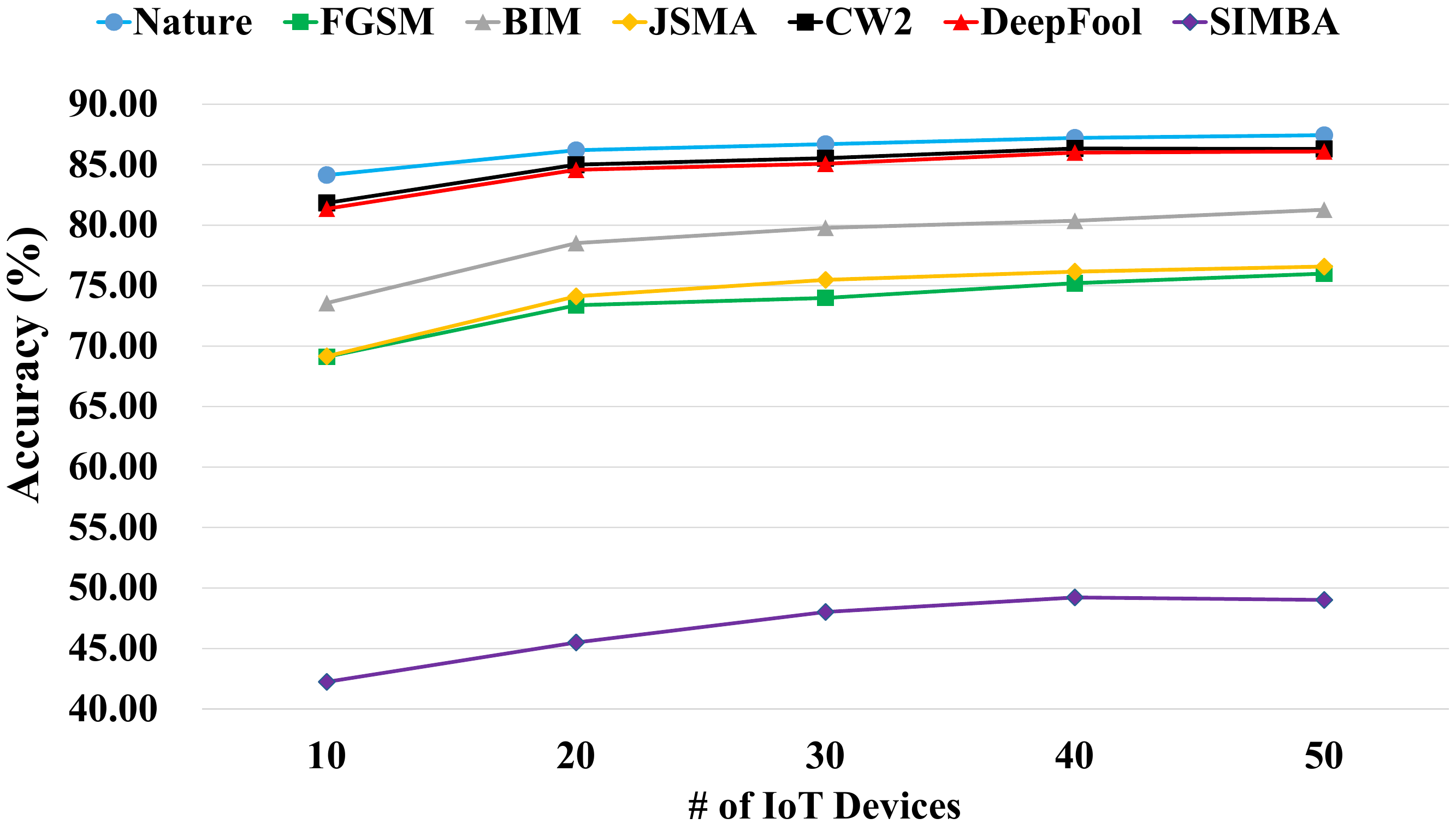}
	\caption{The impact of the number of IIoT devices for our federated defense methods on  
CIFAR10 dataset}
	\label{fig:number-node-cifar10}
\end{figure}

 \figref{fig:number-node-cifar10} shows the results of our federated defense method
on CIFAR10 dataset.  We can observe the similar trend compared with the results shown 
 in \figref{fig:number-node-mnist}.
As an example for the BIM test set, when the number of IIoT devices increases from 10 to 50, 
the accuracy can be improved from 73.5\% to 81.3\%.
Moreover, for the SIMBA test set, the accuracy can be significantly
 improved from 42.2\% to 49.1\%.
In other words, the more devices with high
diversities are involved in federated defense, the more attacks the obtained model can resist.
Therefore, our approach is promising
 especially for large-scale IIoT applications.

\section{Conclusion}\label{sec:conclusions}

Although DNN-based techniques  are becoming popular in 
IIoT applications, they are  suffering 
from an increasing number of adversarial attacks. How to generate  DNNs
that are  immune to 
various types of attacks (especially newly emerging  attacks) is   
becoming a major bottleneck in the deployment of safety-critical IIoT applications.
To address this problem, this paper proposes a novel
federated defense approach for cloud-based IIoT applications.
Based on a modified  
federated learning framework and our proposed  loss function for adversarial learning, 
our approach can effectively synthesize  
DNNs  to accurately
 resist existing adversarial attacks, while the data privacy among different IIoT devices is guaranteed.     
Experimental results on two well-known benchmarks demonstrate
that our approach  
can not only  
improve the overall defense performance against various
 existing adversarial attacks, but also can 
accurately 
detect DNN misbehaviors caused by new kinds of attacks.


\section*{Acknowledgments}

This work received financial support in part from 
National Key Research and Development
 Program of China (Grant \#: 2018YFB2101300), and 
 Natural Science Foundation of China
(Grant \#: 61872147). Mingsong Chen is the corresponding author.

{

}

\end{document}